\documentclass[11pt]{article}

\usepackage[preprint]{acl}

\usepackage{times}
\usepackage{latexsym}

\usepackage[T1]{fontenc}

\usepackage[utf8]{inputenc}

\usepackage{microtype}

\usepackage{inconsolata}

\usepackage{graphicx}

\usepackage{tikz}
\usepackage{todonotes}
\usepackage{soul}
\usepackage{multirow}
\usepackage{tabularx}
\usepackage{float}
\usepackage{amsmath}
\usepackage{multicol}
\usepackage{listings}
\usepackage{listingsutf8}
\usepackage{colortbl}
\usepackage[normalem]{ulem}
\useunder{\uline}{\ul}{}
\usepackage{xcolor}
\usepackage{pifont}

\newcommand{\cmark}{\textcolor{green!60!black}{\ding{51}}}
\newcommand{\xmark}{\textcolor{red}{\ding{55}}}

\title{Commonsense on Demand: Generating and Selectively Integrating Commonsense Knowledge for Natural Language Inference}

\author{
\textbf{Chathuri Jayaweera},
\textbf{Brianna Yanqui},
\textbf{Bonnie J. Dorr}
\\
University of Florida, Gainesville, FL,USA \\
\texttt{\{chathuri.jayawee, yanqui.btheodora, bonniejdorr\}@ufl.edu}}

\begin{document}
\maketitle
\begin{abstract}

Natural Language Inference (NLI) determines whether a premise entails, contradicts, or is neutral with respect to a hypothesis. The task is often framed as emulating human inference, in which commonsense knowledge plays a major role. This study examines whether Large Language Models (LLMs) can reliably generate factual commonsense axioms for NLI, and evaluates their utility on the SNLI and ANLI benchmarks using Llama-3.1-70B and gpt-oss-120b. Because commonsense axioms lack explicit textual references, standard factuality metrics are ill-suited to their evaluation. We therefore introduce a reference-free method using an LLM-as-Judge framework. The evaluation reveals a substantial gap between models: gpt-oss-120b generates predominantly accurate axioms, whereas Llama produces more incorrect than correct ones. We further evaluate three prompting pipelines: direct inference, inference augmented with generated commonsense axioms, and a hybrid approach that selectively incorporates highly factual axioms based on judged factuality. The hybrid approach yields consistent accuracy gains of 3.87\%–8.5\% across tested configurations. Targeted commonsense knowledge also helps models overcome a bias toward the \verb|Neutral| class by providing essential real-world context.

\end{abstract}

\section{Introduction}

Natural Language Inference (NLI), also known as Recognizing Textual Entailment (RTE), is the task of determining, given a sentence pair---a premise (P) and a hypothesis (H)---whether the premise semantically entails the hypothesis (Entailment), contradicts it (Contradiction), or does neither (Neutral). Earlier definitions specify RTE as the directional relationship between P and H, where \textit{Entailment} holds if a typical reader, equipped with general language understanding and background knowledge, would infer that H is likely true upon reading P.


There is little guidance on the amount of common world and language knowledge a system needs to accurately determine the inference relationship between a premise and hypothesis \citep{zaenen_local_2005}. \citet{manning_local_2006} argues that the goal of NLI is to model the inferences made by a typical, informed, attentive reader using everyday 
real-world knowledge. Hence, the task is often framed as emulating human inference, where commonsense knowledge plays a central role.
However, existing models---even those reporting high performance on benchmark datasets---do not consistently learn the correct way to form the connection between the premise and hypothesis \cite{luo_simple_2022, mckenna_sources_2023, liu_were_2023}. A critical and underexplored question, therefore, is whether LLMs can generate the commonsense knowledge needed for correct inference, and whether that knowledge is reliable enough to trust.

\begin{figure}[t]
    \centering
    \small
    \tikzset{every picture/.style={line width=0.75pt}} 

\begin{tikzpicture}[x=0.75pt,y=0.75pt,yscale=-1,xscale=1]

\draw  [draw opacity=0][fill={rgb, 255:red, 214; green, 229; blue, 247 }  ,fill opacity=1 ] (10.4,28.53) .. controls (10.4,15.53) and (20.93,5) .. (33.93,5) -- (277.47,5) .. controls (290.47,5) and (301,15.53) .. (301,28.53) -- (301,52.72) .. controls (301,65.72) and (290.47,76.25) .. (277.47,76.25) -- (33.93,76.25) .. controls (20.93,76.25) and (10.4,65.72) .. (10.4,52.72) -- cycle ;
\draw  [draw opacity=0][fill={rgb, 255:red, 238; green, 245; blue, 231 }  ,fill opacity=1 ] (10,79.33) .. controls (10,63.44) and (22.88,50.57) .. (38.76,50.57) -- (273.24,50.57) .. controls (289.12,50.57) and (302,63.44) .. (302,79.33) -- (302,106.24) .. controls (302,122.12) and (289.12,135) .. (273.24,135) -- (38.76,135) .. controls (22.88,135) and (10,122.12) .. (10,106.24) -- cycle ;
\draw  [draw opacity=0][fill={rgb, 255:red, 249; green, 238; blue, 220 }  ,fill opacity=1 ] (11,45.5) -- (301,45.5) -- (301,78.5) -- (11,78.5) -- cycle ;
\draw  [color={rgb, 255:red, 155; green, 155; blue, 155 }  ,draw opacity=1 ] (11,30.1) .. controls (11,15.68) and (22.68,4) .. (37.1,4) -- (275.3,4) .. controls (289.72,4) and (301.4,15.68) .. (301.4,30.1) -- (301.4,108.39) .. controls (301.4,122.8) and (289.72,134.49) .. (275.3,134.49) -- (37.1,134.49) .. controls (22.68,134.49) and (11,122.8) .. (11,108.39) -- cycle ;

\draw (161.5,105.75) node  [font=\normalsize] [align=left] {\begin{minipage}[lt]{210.12pt}\setlength\topsep{0pt}
\textbf{\textcolor[rgb]{0.49,0.83,0.13}{Commonsense knowledge:}} A person who\\has been shot would typically display signs of\\injury or distress, not a big grin.
\end{minipage}};
\draw (160.35,27.88) node    [font=\normalsize][align=left] {\begin{minipage}[lt]{206.24pt}\setlength\topsep{0pt}
\textbf{\textcolor[rgb]{0.29,0.56,0.89}{Premise:}} A woman with a green headscarf,\\blue shirt and a very big grin.	
\end{minipage}};
\draw (163.5,63.25) node   [font=\normalsize] [align=left] {\begin{minipage}[lt]{210.12pt}\setlength\topsep{0pt}
\textbf{\textcolor[rgb]{0.96,0.65,0.14}{Hypothesis:}} The woman has been shot.
\end{minipage}};

\end{tikzpicture}
    \vspace{-.2in}
    \caption{SNLI instance illustrating the need for commonsense knowledge in inference prediction.}
    \label{fig:ck-example}
    \vspace*{-.2in}
\end{figure}

Commonsense knowledge refers to practical information about how the world works, typically shared by most people \citep{bhargava_commonsense_2022}. In this paper, we refer to \textit{Commonsense Knowledge} as an axiom \citep{zhou_rica_2021}: a basic principle or rule expressed in natural language that reflects widely held understanding and supports everyday inference.
The term ``axiom'' here does not denote a formally defined logical truth (e.g., the Axiom of Choice), but rather a natural language bridging principle used to support inference. These axioms function as self-contained reasoning steps that connect a premise to its hypothesis through background knowledge---for example, inferring that a person standing outside a building is not also inside it. Most adults possess a large number of axioms guiding daily reasoning.


We define an accurate NLI system as one that incorporates implicit axiomatic knowledge to emulate typical human inference. Given the centrality of commonsense reasoning to NLI and related tasks, many efforts have attempted to construct structured commonsense representations. Still, it remains unclear whether existing resources fully capture what is truly needed \citep{gajbhiye_exbert_2021, bauer_ernie-nli_2021}, and whether Large Language Models (LLMs) can serve as a dynamic, on-demand source of commonsense knowledge.


Some NLI instances require reasoning beyond lexical or syntactic alignment. Inferring that a woman with a big grin is most likely not shot (Figure~\ref{fig:ck-example}) involves background knowledge about typical human emotions and expressions---information not present in the text, but accessible through commonsense inference. This raises a dual challenge: not only must a system retrieve or generate relevant commonsense knowledge, but it must also ensure that this knowledge is factually grounded rather than hallucinated.


Recent advances in LLMs, trained on large volumes of human-generated text, suggest they encode significant commonsense knowledge. However, it remains unclear whether LLMs can reliably generate contextually appropriate and factually accurate commonsense axioms for NLI, rather than merely recalling surface-level patterns. Evaluating this capacity requires assessing both whether generated axioms improve downstream NLI predictions and whether they meet independent standards of factuality and relevance. Building on these considerations, we investigate two research questions: (1) Can LLMs reliably generate factual commonsense axioms for NLI premise–hypothesis pairs? (2) Do commonsense axioms improve LLM performance on NLI?


To answer these questions, we design an experimental pipeline that generates and injects commonsense axioms before inference and evaluate both the factuality of generated axioms and their impact on NLI predictions. We further explore selective access to commonsense knowledge conditioned on axiom factuality, and demonstrate performance gains across all datasets and models. Our contributions are summarized as follows:

\begin{itemize}
    \item We introduce a reference-free evaluation method for measuring the factuality of LLM-generated commonsense axioms using an LLM-as-Judge framework.
    
    \item We evaluate the ability of Llama-3.1-70B \cite{grattafiori_llama_2024} and gpt-oss-120b \cite{openai_gpt-oss-120b_2025} to generate commonsense axioms for NLI, assessing the factuality of generated axioms as a measure of LLM commonsense knowledge quality.

    \item We introduce and evaluate a hybrid inference method that selectively incorporates generated commonsense axioms based on judged factuality, demonstrating consistent NLI accuracy improvements across datasets and models.

    \item We show that access to commonsense axioms helps LLMs distinguish \verb|Neutral| from the other inference classes by providing contextual knowledge not available in the literal premise-hypothesis pair.
    
\end{itemize}

\section{Related Work}

Since the introduction of the NLI task, researchers continue to debate what kinds of background knowledge are required. This section reviews efforts to incorporate commonsense and other supporting knowledge into NLI models.

\subsection{Necessity for additional background knowledge in NLI}

Some premise-hypothesis pairs require different types of background knowledge, classified into three categories: linguistic, common, and commonsense \citep{storks_recent_2020}. Linguistic knowledge requires understanding syntactic and semantic structures. Common knowledge refers to widely known facts, often stated explicitly. Commonsense knowledge is the implicit, intuitive understanding that humans use to interpret everyday situations.

\begin{figure*}[h]
  \centering
  \includegraphics[width=1\textwidth]{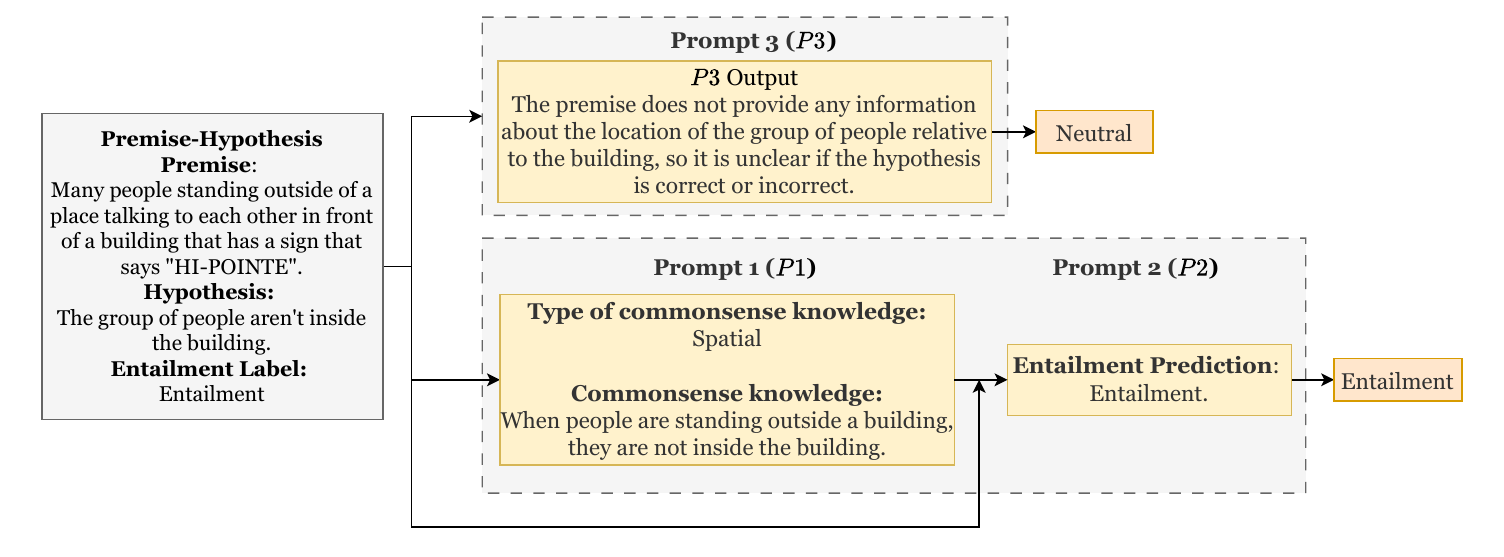}
  \caption{Experiment Methodology}
  \label{fig:ExpMethod}
  \vspace{-.15in}
\end{figure*}

To account for semantic variations between sentence pairs, many NLI systems use external linguistic knowledge resources such as WordNet \cite{miller_wordnet_1995} and hand-crafted inference rules \cite{chen_neural_2018}. However, many implementations also benefit from broader world knowledge, including geographic and spatial understanding, for example, \textit{The cat sat on the mat} implies \textit{The mat was under the cat}. Our study explores whether LLMs can generate suitable commonsense knowledge to meet this requirement.

\subsection{NLI systems with external knowledge}

Several systems explicitly incorporate external commonsense knowledge to improve NLI performance. \citet{camburu_e-snli_2018} investigate whether providing a prior explanation alongside the premise-hypothesis pair improves inference prediction.
Similarly, we
examine how adding commonsense knowledge
impacts prediction accuracy.
\citet{nguyen_effects_2024} extend the e-SNLI dataset by using its highlighted keywords to retrieve relevant commonsense information from ConceptNet \cite{speer_conceptnet_2017} and Google search summaries. However, this approach relies on manually labeled keywords to extract appropriate commonsense information.

ExBERT \cite{gajbhiye_exbert_2021} generates contextual representations that integrate  commonsense knowledge from ConceptNet and Aristo Tuple KGs \cite{dalvi_mishra_domain-targeted_2017}. ERNIE-NLI \cite{bauer_ernie-nli_2021} generates knowledge embeddings using heuristic-based polarity assignments between concept pairs. However, these approaches do not guarantee the relevance of constructed knowledge to the specific premise-hypothesis pair.



\subsection{LLM Factuality Evaluation}

Factuality evaluation has become a central concern as LLMs are known to hallucinate---producing fluent but factually incorrect outputs \citep{ji_survey_2023}. Several metrics have been proposed to address this, including FActScore \citep{min_factscore_2023}, which decomposes generated text into atomic claims verified against retrieved passages; QA-based approaches \citep{wang_factuality_2024} that check answer consistency against a reference; and NLI-based consistency metrics \citep{kryscinski_evaluating_2020, laban_summac_2022} that verify whether generated text is entailed by a source document.

However, these approaches share a critical structural requirement: a trusted reference against which generated text can be verified. This works for summarization and factoid QA, where source documents or gold answers exist, but breaks down for commonsense statements. Since commonsense knowledge is implicit and rarely documented explicitly \citep{bhargava_commonsense_2022}, standard reference-based pipelines cannot reliably verify commonsense axioms, leaving existing metrics ill-suited for this setting.

To address this, we modify the factuality framework of \citet{zheng_large_2024} using an LLM-as-Judge implementation \citep{zheng_judging_2023}, enabling reference-free evaluation of commonsense axiom quality in our open-ended generation setting.


\section{RQ1: Can LLMs reliably generate factual commonsense axioms for NLI premise–hypothesis pairs?}

For RQ1, we prompt two LLMs on two NLI datasets to generate commonsense axioms. The generated axioms are then evaluated for factuality. 


\subsection{Datasets}

We use two established English Natural Language Inference (NLI) datasets: the Stanford Natural Language Inference (SNLI) dataset \cite{bowman_large_2015} and the Adversarial NLI (ANLI) dataset \cite{nie_adversarial_2020}. SNLI consists of premise–hypothesis pairs derived from short image captions, representing concise, everyday situations with a relatively simple structure. ANLI contains longer, multi-sentence premises from diverse domains and was created through an adversarial human-and-model-in-the-loop process to produce more challenging inference problems. Together, these datasets provide diversity in linguistic complexity, context length, and difficulty, enabling evaluation across different NLI settings while maintaining controlled experimental conditions. 

Even though both datasets are well-studied and report high baseline scores, it is not guaranteed that the model has correctly learned the reasoning relationship between the premise and hypothesis \citep{poliak_hypothesis_2018, gururangan_annotation_2018, luo_simple_2022}. Therefore, our experiments ground the prediction on the related commonsense axioms, thereby promoting a more reliable reasoning process. From each dataset, we sample 2000 annotated sentence pairs spanning all three inference labels. The data distribution over three inference classes for both datasets is shown in Table \ref{tab:data-stats}.

\begin{table}[]
\centering
\footnotesize
\begin{tabular}{lllll}
\hline
Dataset & Entailment & Contradiction & Neutral & Total \\ \hline
SNLI    & 689        & 651           & 660     & 2000  \\
ANLI    & 771        & 585           & 644     & 2000  \\ \hline
\end{tabular}%
\caption{Dataset statistics over the inference classes.}
\label{tab:data-stats}
\vspace{-.2in}
\end{table}

\subsection{Models}


We select two open-source LLMs from different families to ensure diversity in model architectures and reasoning behavior: Llama-3.1-70b-Instruct \cite{grattafiori_llama_2024} and gpt-oss-120b \cite{openai_gpt-oss-120b_2025}. This selection allows us to examine the impact of commonsense knowledge injection across qualitatively different LLM families while maintaining controlled experimental conditions. For each premise-hypothesis pair, we use three prompt types to either generate or add commonsense axioms before predicting inference labels, or predict them without added axioms.

\begin{figure}[ht]
\vspace{-.1in}
    \centering
    \small
    \tikzset{every picture/.style={line width=0.75pt}} 

\begin{tikzpicture}[x=0.75pt,y=0.75pt,yscale=-1,xscale=1]

\draw    (75,82.5) -- (349,82.5) ;
\draw    (75,141.5) -- (351,141.5) ;

\draw (205,57.25) node   [align=left] {\begin{minipage}[lt]{191.76pt}\setlength\topsep{0pt}
{\fontfamily{ptm}\selectfont {\footnotesize Provide the commonsense knowledge that is necessary to understand the relationship between a given \textcolor[rgb]{0.29,0.56,0.89}{\textbf{premise}} and a \textbf{\textcolor[rgb]{0.96,0.65,0.14}{hypothesis}}. Focus on concrete, factual knowledge that is widely accepted. }} \\
\end{minipage}};
\draw (63.5,49.25) node  [rotate=-270] [align=left] {\begin{minipage}[lt]{42.84pt}\setlength\topsep{0pt}
{\fontfamily{ptm}\selectfont {\footnotesize \textbf{Instruction}}}
\end{minipage}};
\draw (204.5,113.25) node   [align=left] {\begin{minipage}[lt]{187pt}\setlength\topsep{0pt}
{\fontfamily{ptm}\selectfont {\footnotesize \textcolor[rgb]{0.29,0.56,0.89}{\textbf{Premise}}: \{premise\}}}\\{\fontfamily{ptm}\selectfont {\footnotesize \textcolor[rgb]{0.96,0.65,0.14}{\textbf{Hypothesis}}: \{hypothesis\}}}\\{\fontfamily{ptm}\selectfont {\footnotesize Type of commonsense knowledge: \{type\}}}\\{\fontfamily{ptm}\selectfont {\footnotesize Commonsense knowledge: \{commonsense-axiom\}}}
\end{minipage}};
\draw (65.5,111.25) node  [rotate=-270] [align=left] {\begin{minipage}[lt]{42.84pt}\setlength\topsep{0pt}
{\fontfamily{ptm}\selectfont {\footnotesize \textbf{Examples}}}
\end{minipage}};

\end{tikzpicture}
    \vspace*{-.2in}

    \caption{\(P1\) prompt format: We provide three in-context examples with the instructions for \(P1\) prompt and provide the test premise-hypothesis pair at the end.}
    \label{fig:prompt1}
    \vspace{-.23in}
\end{figure}

\subsection{Methodology}

We use \(P1\) to generate targeted commonsense axioms for each premise-hypothesis pair. \(P1\) is run five times per premise-hypothesis pair. The generated axioms are evaluated for factuality (Section~\ref{sec:factuality-eval}), and results are averaged to reduce variance. 

\noindent\textbf{Commonsense axiom generation (\(P1\)):} As shown in Figure~\ref{fig:ExpMethod}, \(P1\) is used to identify the commonsense knowledge type required for a premise-hypothesis pair and generate a commonsense axiom (See Figure \ref{fig:prompt1}) to ensure the relevance of the generated knowledge to the specified premise-hypothesis pair. This design prioritizes precision over breadth. Future work will explore multi-sentence axioms to assess whether additional context improves inference quality.

\subsection{Factuality Evaluation}
\label{sec:factuality-eval}

We adapt the factuality metric of \citet{zheng_large_2024} for the NLI task. Unlike question answering, NLI provides no definitive ground truth for evaluating generated commonsense axioms. We therefore rely on automatic annotations.

We use Claude-3.5-Sonnet\footnote{\url{https://www.anthropic.com/news/claude-3-5-sonnet}} as an LLM judge to rate the helpfulness of each generated axiom for inference prediction on a 1–10 scale (See Appendix \ref{sec:appendix-helpfulness}). We provide the premise \(T_{p_i}\), hypothesis \(T_{h_i}\), commonsense axiom \(CK_i\) and gold inference label  \(l_i\) to prompt \(J_{h}\), returning a score \(r_{h_i}\) between 1 and 10 (Eq. \ref{Eq:help-rating}). Scores $\geq 9$ are converted to binary labels \(h_i\) $=1$, and all others to $0$ (Eq. \ref{Eq:threshold1}).
\vspace{-0.06in}
\begin{align}
\label{Eq:help-rating}
 J_{h}(T_{p_i}, T_{h_i}, CK_i, l_i) = r_{h_i}\\
 h_i = 
 \begin{cases}
 1, & r_{h_i} \geq 9 \\
 0, & r_{h_i} < 9 
 \label{Eq:threshold1}
 \end{cases}
\end{align}

Following these steps, we use the LLM-as-Judge model to annotate the \(P1\)-generated commonsense axioms and calculate the Net Correct Rate (NCR) factuality metric as defined by \citet{zheng_large_2024}, using binary labels (Eqs. \ref{Eq:Ncorrect}--\ref{Eq:Nwrong}). The LLM judge evaluates the factual plausibility of each generated axiom independently of downstream prediction outcomes. Table \ref{tab:factuality-eval} shows the factuality assessment for \(P1\)-generated commonsense axioms.
\vspace{-0.1in}
\begin{align}
\label{Eq:Ncorrect}
    CR = \frac{\sum_{i=1}^{N} [h_i=1]}{N}\\
    WR = \frac{\sum_{i=1}^{N} [h_i=0]}{N}
    \label{Eq:Nwrong}
\end{align}

\subsection{ Results and Analysis}

Table~\ref{tab:factuality-eval} reports the factuality of commonsense axioms generated using each model across both datasets, measured by Correct Rate (CR), Wrong Rate (WR), and Net Correct Rate (NCR = CR-WR). The results reveal a substantial gap between the two models. gpt-oss-120b achieves positive NCR scores of 27.94 and 14.34 on SNLI and ANLI, respectively, indicating that it generates more accurate axioms than inaccurate ones across both datasets. Llama-3.1-70b-Instruct, by contrast, yields negative NCR scores of -12.07 and -24.96, meaning it generates more incorrect than correct axioms overall. Consequently, gpt-oss-120b appears to be more reliable for factual commonsense knowledge generation than Llama-3.1-70b-Instruct. Both models show degraded factuality on ANLI relative to SNLI, consistent with ANLI's adversarial construction and greater inference difficulty.

\begin{table*}[ht]
\centering
\begin{tabular}{lcccc}
\hline
\multirow{2}{*}{Factuality Metric} & \multicolumn{2}{l}{Llama-3.1-70b-Instruct} & \multicolumn{2}{l}{gpt-oss-120b} \\ \cline{2-5} 
                               & SNLI   & ANLI   & SNLI  & ANLI  \\ \hline
Correct Rate (CR)\%              & 43.96  & 37.51  & 63.97 & 57.17 \\
Wrong Rate (WR)\%                & 56.04  & 62.48  & 36.02 & 42.82 \\
Net Correct Rate (NCR = CR-WR)\% & -12.07 & -24.96 & 27.94 & 14.34 \\ \hline
\end{tabular}%
\caption{Factuality evaluation for \(P1\)-generated commonsense axioms.}
\label{tab:factuality-eval}
\end{table*}


\section{RQ2: Do commonsense axioms improve LLM performance on NLI?}

We use two prompts, \(P2\) and \(P3\), to predict the inference label for each premise-hypothesis pair in both datasets. In \(P2\), the axioms generated by \(P1\) are appended to provide targeted commonsense knowledge before prediction, whereas \(P3\) predicts the inference label without access to such axioms. Figure~\ref{fig:ExpMethod} summarizes the experimental workflow and illustrates an example premise–hypothesis pair producing \verb|Entailment| and \verb|Neutral| predictions. 

\begin{figure}[h]
\vspace{-.05in}
    \centering
    \small
    \tikzset{every picture/.style={line width=0.75pt}} 

\begin{tikzpicture}[x=0.75pt,y=0.75pt,yscale=-1,xscale=1]

\draw    (11,151.5) -- (284,151.5) ;
\draw    (11,213.5) -- (284,213.5) ;

\draw (142.88,88.5) node   [align=left] {\begin{minipage}[lt]{190.23pt}\setlength\topsep{0pt}
{\fontfamily{ptm}\selectfont {\footnotesize Based on the provided \textcolor[rgb]{0.29,0.56,0.89}{\textbf{premise}}, \textbf{\textcolor[rgb]{0.96,0.65,0.14}{hypothesis}}, and \textbf{\textcolor[rgb]{0.49,0.83,0.13}{commonsense knowledge}}, predict the textual entailment relationship between the context and the statement. Respond with one of the following labels:}}\\{\fontfamily{ptm}\selectfont {\footnotesize \textbf{\textcolor[rgb]{0.25,0.46,0.02}{Entailment}} (The premise definitely supports the hypothesis)}}\\{\fontfamily{ptm}\selectfont {\footnotesize \textbf{\textcolor[rgb]{0.82,0.01,0.11}{Contradiction}} (The premise definitely contradicts the hypothesis)}}\\{\fontfamily{ptm}\selectfont {\footnotesize \textbf{\textcolor[rgb]{0.97,0.91,0.11}{Neutral}} (The premise neither supports nor contradicts the hypothesis)}}\\
\end{minipage}};
\draw (-0.5,80.25) node  [rotate=-270] [align=left] {\begin{minipage}[lt]{42.84pt}\setlength\topsep{0pt}
{\fontfamily{ptm}\selectfont {\footnotesize \textbf{Instruction}}}
\end{minipage}};
\draw (138.5,184.25) node   [align=left] {\begin{minipage}[lt]{187pt}\setlength\topsep{0pt}
{\fontfamily{ptm}\selectfont {\footnotesize \textcolor[rgb]{0.29,0.56,0.89}{\textbf{Premise}}: \{premise\}}}\\{\fontfamily{ptm}\selectfont {\footnotesize \textcolor[rgb]{0.96,0.65,0.14}{\textbf{Hypothesis}}: \{hypothesis\}}}\\{\fontfamily{ptm}\selectfont {\footnotesize \textbf{\textcolor[rgb]{0.49,0.83,0.13}{Commonsense Knowledge}}: \{commonsense-axiom\}}}\\{\fontfamily{ptm}\selectfont {\footnotesize Output: \{entailment-relationship\}}}
\end{minipage}};
\draw (-0.5,173.25) node  [rotate=-270] [align=left] {\begin{minipage}[lt]{42.84pt}\setlength\topsep{0pt}
{\fontfamily{ptm}\selectfont {\footnotesize \textbf{Examples}}}
\end{minipage}};

\end{tikzpicture}
    \vspace*{-.15in}

    \caption{\(P2\) prompt format: We provide three in-context examples with the instructions for \(P2\) prompt and provide the test premise-hypothesis pair and commonsense axiom from \(P1\) at the end.}
    \vspace{-.13in}
    \label{fig:prompt2}
\end{figure}

\begin{figure}[h]
    \centering
    \small
    \tikzset{every picture/.style={line width=0.75pt}} 

\begin{tikzpicture}[x=0.75pt,y=0.75pt,yscale=-1,xscale=1]

\draw    (8,180.5) -- (275,180.5) ;
\draw    (9,230.5) -- (276,230.5) ;

\draw (143,116.5) node   [align=left] {\begin{minipage}[lt]{191.76pt}\setlength\topsep{0pt}
{\fontfamily{ptm}\selectfont {\footnotesize Consider the \textbf{\textcolor[rgb]{0.29,0.56,0.89}{premise}} and the \textbf{\textcolor[rgb]{0.96,0.65,0.14}{hypothesis}} provided below.}}\\{\fontfamily{ptm}\selectfont {\footnotesize \textbf{\textcolor[rgb]{0.29,0.56,0.89}{Premise}}: \{premise\}}}\\{\fontfamily{ptm}\selectfont {\footnotesize \textbf{\textcolor[rgb]{0.96,0.65,0.14}{Hypothesis}}: \{hypothesis\}}}\\{\fontfamily{ptm}\selectfont {\footnotesize Choose the appropriate textual entailment label for the given premise and hypothesis:}}\\{\fontfamily{ptm}\selectfont {\footnotesize Definitely correct (\textbf{\textcolor[rgb]{0.25,0.46,0.02}{Entailment}}); or Definitely incorrect (\textbf{\textcolor[rgb]{0.82,0.01,0.11}{Contradiction}}); or Neither (\textbf{\textcolor[rgb]{0.97,0.91,0.11}{Neutral}}).}}\\{\fontfamily{ptm}\selectfont {\footnotesize Format the response as: The label selection: Brief one sentence explanation.}}\\
\end{minipage}};
\draw (3.5,112.25) node  [rotate=-270] [align=left] {\begin{minipage}[lt]{42.84pt}\setlength\topsep{0pt}
{\fontfamily{ptm}\selectfont {\footnotesize \textbf{Instruction}}}
\end{minipage}};
\draw (141.5,205.63) node   [align=left] {\begin{minipage}[lt]{187pt}\setlength\topsep{0pt}
{\fontfamily{ptm}\selectfont {\footnotesize \textcolor[rgb]{0.29,0.56,0.89}{\textbf{Premise}}: \{premise\}}}\\{\fontfamily{ptm}\selectfont {\footnotesize \textcolor[rgb]{0.96,0.65,0.14}{\textbf{Hypothesis}}: \{hypothesis\}}}\\{\fontfamily{ptm}\selectfont {\footnotesize Output: \{entailment-relationship\}}}
\end{minipage}};
\draw (4,199.25) node  [rotate=-270] [align=left] {\begin{minipage}[lt]{42.84pt}\setlength\topsep{0pt}
{\fontfamily{ptm}\selectfont {\footnotesize \textbf{Examples}}}
\end{minipage}};

\end{tikzpicture}
    \vspace*{-.1in}

    \caption{\(P3\) prompt format: We provide the instructions and three in-context examples with the test premise-hypothesis pair without adding a commonsense axiom to obtain the P3 inference class prediction.}
    \vspace{-.17in}
    \label{fig:prompt3}
\end{figure}

\noindent\textbf{Inference prediction with generated axioms (\(P2\)):} Figure~\ref{fig:ExpMethod} shows an instance where using commonsense axioms results in a correct prediction. For this, we use \(P2\), which is executed in a separate session and takes as input the commonsense axiom generated from \(P1\). The axiom is then appended to the premise-hypothesis pair during inference (see Figure \ref{fig:prompt2}). Each prompt is processed in an independent execution context, with no retained memory between runs. The term `prompt' is used in a modular sense to reflect this isolation, distinguishing between distinct stages in the reasoning pipeline. We use \(P2\) responses to analyze how access to generated knowledge affects the model's predictions (entailment, contradiction, or neutral). 

\noindent\textbf{Direct inference prediction with explanation (\(P3\)):} Figure~\ref{fig:ExpMethod} shows an instance where the inference label is mispredicted in the absence of a commonsense axiom. \(P3\) serves as a comparison to \(P1\) and \(P2\). It asks the model to predict the inference label without adding external commonsense axioms, but requires a short explanation (See Figure \ref{fig:prompt3}). We treat the explanation as the model's implicit reasoning. The results obtained from \(P3\) serve as a baseline to evaluate the impact of integrating the generated axioms.

\subsection{Hybrid experiment with selective access to commonsense knowledge}

We implement a hybrid approach in which access to generated commonsense knowledge is controlled by a helpfulness rating, allowing us to evaluate whether selectively adding such knowledge improves NLI performance beyond the P3 pipeline, where no external commonsense knowledge is available during inference.

This experiment tests the hypothesis that if an oracle could determine whether a premise–hypothesis pair requires commonsense knowledge, selectively providing such knowledge would improve prediction accuracy. To approximate such an oracle, we use the helpfulness rating to filter instances that require additional knowledge.

\begin{table*}[h]
\centering
\resizebox{\textwidth}{!}{%
\begin{tabular}{lcccccc}
\hline
\multirow{3}{*}{Inference Type} & \multicolumn{6}{c}{Model (Dataset: SNLI)}                                                                             \\ \cline{2-7} 
                                & \multicolumn{3}{c}{Llama-3.1-70b-Instruct}                & \multicolumn{3}{c}{gpt-oss-120b}                          \\ \cline{2-7} 
 &
  \multicolumn{1}{c}{P1+P2} &
  \multicolumn{1}{c}{P3} &
  \multicolumn{1}{c}{Hybrid (\(\theta=9\))} &
  \multicolumn{1}{c}{P1+P2} &
  \multicolumn{1}{c}{P3} &
  \multicolumn{1}{c}{Hybrid (\(\theta=9\))} \\ \hline
Entailment                      & {\ul 98.32 (0.22)} & 96.00 (0.36) & 96.95 (0.26)          & {\ul 93.52 (0.82)} & 86.32 (0.22) & 92.19 (0.57)          \\
Contradiction                   & {\ul 65.59 (1.49)} & 53.41 (0.82) & 64.52 (1.34)          & {\ul 69.01 (0.65)} & 47.53 (0.85) & 68.43 (0.60)          \\
Neutral                         & 31.53 (0.62)       & 57.13 (0.76) & 56.97 (0.58)          & 73.86 (1.37)       & 95.78 (0.61) & 94.80 (0.62)          \\ \hline
Overall                         & 65.94 (0.37)       & 69.53 (0.27) & \textbf{73.41 (0.59)} & {\ul 79.17 (0.74)} & 76.84 (0.38) & \textbf{85.34 (0.45)} \\ \hline
\end{tabular}%
}
\caption{Average prediction accuracy for the SNLI dataset for each inference class and overall, both with \((P1+P2)\) and without \((P3)\) pre-generated commonsense axioms, as well as in the hybrid setting. Standard deviations are provided in parentheses. Cases where \(P1+P2\) surpasses \(P3\) performance are underlined, while cases where the hybrid setting achieves the highest performance are in bold.}
\label{tab:snli-pred-error-analysiz}
\vspace{-.15in}

\end{table*}

\begin{table*}[h]
\centering
\resizebox{\textwidth}{!}{%
\begin{tabular}{lllllll}
\hline
\multirow{3}{*}{Inference Type} & \multicolumn{6}{c}{Model (Dataset: ANLI)}                                                                             \\ \cline{2-7} 
                                & \multicolumn{3}{c}{Llama-3.1-70b-Instruct}                & \multicolumn{3}{c}{gpt-oss-120b}                          \\ \cline{2-7} 
 &
  \multicolumn{1}{c}{P1+P2} &
  \multicolumn{1}{c}{P3} &
  \multicolumn{1}{c}{Hybrid (\(\theta=8\))} &
  \multicolumn{1}{c}{P1+P2} &
  \multicolumn{1}{c}{P3} &
  \multicolumn{1}{c}{Hybrid (\(\theta=9\))} \\ \hline
Entailment                      & {\ul 85.37 (0.80)} & 73.31 (0.75) & 81.87 (0.73)          & {\ul 80.20 (0.50)} & 65.75 (0.95) & 74.87 (0.49)          \\
Contradiction                   & {\ul 65.81 (1.95)} & 63.34 (1.19) & \textbf{68.55 (1.85)} & {\ul 76.04 (0.95)} & 63.90 (0.81) & \textbf{76.18 (0.83)} \\
Neutral                         & 36.91 (0.22)       & 51.39 (0.77) & 48.60 (0.60)          & 42.96 (1.27)       & 76.06 (0.95) & 69.45 (0.98)          \\ \hline
Overall                         & {\ul 63.84 (0.41)} & 63.25 (0.65) & \textbf{67.12 (0.52)} & {\ul 66.77 (0.71)} & 68.61 (0.21) & \textbf{73.46 (0.38)} \\ \hline
\end{tabular}%
}
\caption{Average prediction accuracy for the ANLI dataset for each inference class and overall, both with \((P1+P2)\) and without \((P3)\) pre-generated commonsense axioms, as well as in the hybrid setting. Standard deviations are provided in parentheses. Cases where \(P1+P2\) surpasses \(P3\) performance are underlined, while cases where the hybrid setting achieves the highest performance are in bold.}
\label{tab:anli-pred-error-analysiz}
\vspace{-.15in}
\end{table*}

We use the helpfulness ratings obtained in Section~\ref{sec:factuality-eval} for this implementation. To identify the optimal filtering threshold, 
we sample 500 premise-hypothesis pairs from the 2000 instances in SNLI and ANLI. For each threshold value \(\theta\) in the range \([0, 1, 2, \dots, 10]\), we evaluate the helpfulness rating \(r_{h_i}\) of the generated axiom \(CK_{i}\) for premise-hypothesis pair \((T_{p_i}, T_{h_i})\). 
If 
\(r_{h_i}\geq \theta\), 
the prediction from \(P1 + P2\) is selected; otherwise, the prediction from \(P3\) is used. This process is repeated across all five commonsense-axiom runs.

Predictions from the hybrid process are evaluated against gold labels to identify thresholds that maximize performance. Accuracies are averaged across the five runs to obtain more stable estimates. The variation in overall accuracy and label-wise accuracy across threshold values is shown in Appendix~\ref{sec:appendix-acc-variation}. The highest overall performance is achieved at a helpfulness threshold of 8 for ANLI and 9 for SNLI with the Llama-3.1-70b-Instruct model and a threshold of 9 for both datasets with gpt-oss-120b model. Accordingly, dataset and model combination-specific threshold values are applied to the remaining 1500 premise-hypothesis pairs to assess whether these observations generalize across the full dataset. 

\subsection{Prediction Error Analysis} \label{sec:eval}

We evaluate responses from \(P1+P2\), \(P3\), and the hybrid experiment in terms of a prediction error analysis. The prediction error analysis examines how those axioms affect accuracy by classifying prediction errors based on comparisons between the model's inference label predictions from \(P1+P2\), \(P3\), and the hybrid pipeline and the gold label.

We define six distinct error types based on gold-prediction label mismatches by enumerating each possible mismatch between the gold inference label and the model’s predicted label. We count the occurrences of these error types and report their average frequency and standard deviation across five runs to assess which approach, \(P1+P2\), \(P3\), or the hybrid experiment, more effectively reduces prediction errors. 

\subsection{Results and Analysis}
This section provides a comprehensive analysis of the obtained evaluation results. 

\subsection{Prediction Error Analysis}

\begin{table*}[h]
\centering
\resizebox{\textwidth}{!}{%
\begin{tabular}{lll}
\hline
Dataset &
  Premise (P) + Hypothesis (H) &
  Axioms and Predictions \\ \hline
SNLI &
  \begin{tabular}[c]{@{}l@{}}P: A person in orange clothing rests above a metro entrance.\\ H: Someone is standing near a metro station.\end{tabular} &
  \begin{tabular}[c]{@{}l@{}}P1: ``A person resting near or above a metro\\ entrance is likely to be near a metro station''\\ P1+P2: Entailment \cmark\\ \\ P3: ``The statement is possible but does not\\ necessarily follow from the given context,\\ which specifies the person is above the metro\\ entrance, not near it''\\ P3: Neutral \xmark\end{tabular} \\ \cline{2-3} 
 &
  \begin{tabular}[c]{@{}l@{}}P: A woman, standing behind a girl, helping the girl with an experiment.\\ H: A mother is helping her child complete the experiment.\end{tabular} &
  \begin{tabular}[c]{@{}l@{}}P1: ``A mother is typically a woman who helps her\\ child, often in a supportive or guiding role''\\ P1+P2: Entailment \xmark\\ \\ P3: ``The statement assumes the relationship between\\ the two individuals is that of a mother and child,\\ but the context does not explicitly confirm this''\\ P3: Neutral \cmark\end{tabular} \\ \hline
ANLI &
  \begin{tabular}[c]{@{}l@{}}P: Maurice Généreux is a Canadian physician who was\\ convicted in 1998 of prescribing medications to two HIV\\ positive men in Toronto, Canada in 1996; medications that\\ subsequently allowed the men, Mark Jewitt and Aaron\\ Mcginn, to commit suicide in 1996. Généreux was the first\\ doctor in North America to be convicted of assisting a suicide\\ (followed in 1999 by Jack Kevorkian).\\ H: Maurice Généreux was influenced by the actions of Jack Kevorkian.\end{tabular} &
  \begin{tabular}[c]{@{}l@{}}P1: ``A person cannot be influenced by the\\ actions of another person if those actions\\ occurred after their own actions took place''\\ P1+P2: Contradiction \cmark\\ \\ P3: ``The statement cannot be confirmed or\\ denied based on the given context, as it only\\ mentions the order of convictions but does\\ not provide information about the influence of\\ Jack Kevorkian's actions on Maurice Généreux''\\ P3: Neutral \xmark\end{tabular} \\ \cline{2-3} 
 &
  \begin{tabular}[c]{@{}l@{}}P: How to obtain a copy of your birth certificate in connecticut\\ <br>Determine if you can request a birth certificate. If you want\\ to request a birth certificate that is over 100 years old, you only\\ have to be 18 to request it. If the birth certificate is less than 100\\ years old and does not belong to you, you have to fall under one of the\\ following categories : [substeps] Immediate family, which includes\\ parent, guardian, grandparents, or spouse.\\ H: Immediate family includes siblings.\end{tabular} &
  \begin{tabular}[c]{@{}l@{}}P1: ``Siblings are typically considered part of\\ an individual's immediate family, which may or\\ may not be included in specific definitions or \\ policies''\\ P1+P2: Neutral \xmark\\ \\ P3: ``The context explicitly lists the members of\\ immediate family and does not include siblings''\\ P3: Contradiction \cmark\end{tabular} \\ \hline
\end{tabular}%
}
\caption{Examples illustrating when access to commonsense axioms is beneficial and when it can be detrimental}
\label{tab:qualitative-analysis}
\vspace{-0.2in}
\end{table*}

Prediction error analyses in Tables~\ref{tab:snli-pred-error-analysiz}--\ref{tab:anli-pred-error-analysiz} show that, across the two datasets and models, adding commonsense axioms prior to inference prediction improves the models' ability to accurately distinguish \verb|Entailment| from the other two inference classes. In SNLI, it also enhances the identification of \verb|Contradiction| relationships. In ANLI, the performance gains shift between \verb|Contradiction| and \verb|Neutral| across the models with prior access to commonsense axioms. However, the overall accuracy of the \(P1+P2\) pipeline surpasses \(P3\) only on the SNLI dataset with gpt-oss-120b and the ANLI dataset with Llama-3.1-70b-Instruct.

Notably, the hybrid approach---where access to commonsense axioms is determined by helpfulness---yields the highest overall accuracy across all datasets and model configurations, with 
improvements ranging from 3.87\% to 8.5\%. These results demonstrate that incorporating high-quality commonsense knowledge during inference prediction is particularly effective compared to the \(P1+P2\) and \(P3\) pipelines.

Furthermore, selectively accessing commonsense axioms combines the strengths of the \(P1+P2\) and \(P3\) pipelines, preserving strong performance for inference classes where each pipeline excels individually. For example, in ANLI, the hybrid approach maintains the high \verb|Entailment| accuracy of \(P1+P2\) while improving performance on the remaining classes.

These results suggest that commonsense knowledge is most beneficial when applied selectively, ensuring that highly factual axioms are used only when needed. Although helpfulness ratings serve as a filtering criterion in this experiment, such signals are typically unavailable in real-world settings. Accordingly, further exploration is required to identify cases that warrant additional knowledge.


\subsection{Qualitative Analysis}

We conduct a qualitative analysis to identify cases where access to commonsense axioms improves performance over the \(P3\) pipeline. Table~\ref{tab:qualitative-analysis} presents representative instances from SNLI and ANLI datasets where access to commonsense axioms leads to correct predictions.

In the first examples from both SNLI and ANLI, comparing the \(P1\)-generated axioms with the \(P3\) explanations shows that the \(P3\) pipeline relies on literal reasoning and often fails to infer relationships that are not explicitly stated. In contrast, the \(P1\)-generated axioms leverage real-world knowledge to identify the relationship between the premise and hypothesis. These examples illustrate how spatial and temporal commonsense knowledge enables correct inference. 

Most cases where \(P1\)-generated axioms improve predictions occur when the \(P3\) pipeline returns a \verb|Neutral| prediction. This suggests that both models default to the \verb|Neutral| class when additional contextual knowledge is unavailable, revealing a bias towards \verb|Neutral| class. Providing additional context through commonsense axioms helps guide the models toward the correct inference label. 

By contrast, the second examples from SNLI and ANLI show cases where overly generalized axioms generated by \(P1\) lead to incorrect predictions, while the more literal reasoning of \(P3\) yields the correct inference label. This suggests that although LLMs often encode useful commonsense knowledge, their ability to apply it appropriately remains limited (e.g., a woman helping a girl does not necessarily imply a mother-child relationship).

\section{Conclusion}

In this work, we investigate two questions: whether LLMs can reliably generate factually accurate commonsense axioms for NLI, and whether incorporating such axioms improves LLM inference accuracy. To address these questions, we evaluated Llama-3.1-70B and gpt-oss-120b on the SNLI and ANLI datasets across three settings: without axioms, with axioms, and in a hybrid setting that selectively incorporates axioms based on their factuality.

On the factuality dimension, our results reveal a substantial gap between the two models. gpt-oss-120b generates a majority of accurate axioms across both datasets, whereas Llama-3.1-70b-Instruct produces more incorrect axioms than correct ones, particularly on the more demanding ANLI dataset. This finding shows that LLMs vary considerably in their capacity to serve as reliable generators of commonsense knowledge, and that factuality cannot be assumed without explicit evaluation. Furthermore, both models produce less factual axioms on ANLI than on SNLI, suggesting that adversarial inference contexts elicit less reliable reasoning from LLMs.
\vspace{-0.04in}

On the NLI task, the hybrid approach---selectively providing commonsense axioms based on judged helpfulness---yields the largest performance gains, ranging from 3.87\% to 8.5\%. Qualitative analysis reveals that axioms are particularly effective at helping models distinguish \verb|Entailment| and \verb|Contradiction| cases, and at mitigating the bias toward the \verb|Neutral| class. However, overly generalized axioms may introduce errors. Hence, the value of generated axioms depends on their factual accuracy. 

\vspace{-0.0315in}

Taken together, these findings suggest that LLMs can generate useful commonsense knowledge for NLI, but only when the generated axioms are sufficiently factual. Two key challenges remain: developing more reliable commonsense generation techniques---especially for smaller or weaker models---and identifying, without oracle signals, which inference instances benefit from external axiomatic knowledge.




\section{Future Work}

Future research will aim to improve the accuracy and reliability of commonsense axiom generation for NLI, with a focus on guiding LLMs to produce more factual and consistent axioms. We will further explore early filtering methods to identify inference cases most likely to benefit from external commonsense knowledge, such as ambiguous premise–hypothesis pairs \cite{jiang_investigating_2022, liu_were_2023, jayaweera_disagreement_2025}.


Evaluating a broader range of LLMs and their ability to generate high-quality commonsense knowledge, as well as developing new resources for commonsense reasoning, will further advance this work. These efforts will help create more robust and interpretable NLI systems that better approximate human inference.

\section*{Limitations}

This study examines the use of LLMs to generate commonsense axioms for premise-hypothesis pairs to assist in identifying the correct inference relationship. While we evaluate this approach empirically, several limitations remain.

First, our evaluation relies on an LLM to assess the helpfulness of generated commonsense axioms without full human annotation. Although we manually review a small subset of outputs, the overall accuracy of LLM-generated ratings is not comprehensively validated. Nonetheless, our initial reviews suggest that the assessments are generally reliable.

Second, we evaluate the reliability and impact of LLM-generated commonsense axioms using only two models. While the results demonstrate potential, testing across a broader range of LLMs would provide a more robust validation.


Finally, our study is limited to the English language. All findings are validated solely in English, which narrows the generalizability of our conclusions. We aim for this work to serve as a foundation for future research in multilingual NLI settings.

\section*{Ethics Statement}

Our study uses LLM-generated commonsense axioms in its experiments. Some examples reveal a tendency toward overgeneralization, which may reflect harmful assumptions. A manual review of a small sample suggests that these overgeneralized axioms are typically rated low in factuality, leading to incorrect predictions. Nonetheless, we do not conduct a full audit of potential harms and acknowledge this is an area for future examination. 

We use publicly available datasets---SNLI and ANLI---under CC BY-SA 4.0 and CC BY-NC 4.0 licenses, respectively. 
We ensure proper attribution and rely on the datasets' original anonymization and ethical review procedures. Although we do not explicitly filter generated axioms for personally identifiable information (PII), the prompts are designed to elicit generalized knowledge rather than personal content, making PII inclusion highly unlikely. The prompt details are included in the study. 

We use the Llama-3.1-70b-Instruct and gpt-oss-120b models to generate axioms and the Claude-3.5-Sonnet model to evaluate them, both accessed via institution-internal API (not third party) in inference-only mode. No models were trained as a part of this study. The total cost of API usage for reported experiments was under \$150.

We used AI assistance for stylistic edits (e.g., paraphrasing or synonym suggestions), but all substantive content was authored by the research team.

\bibliography{references}

\newpage
\appendix

\section{Helpfulness Evaluation Prompt} \label{sec:appendix-helpfulness}

\begin{figure}[h]
    \centering
    \small
    \tikzset{every picture/.style={line width=0.75pt}} 

\begin{tikzpicture}[x=0.75pt,y=0.75pt,yscale=-1,xscale=1]

\draw    (13,251.5) -- (280,251.5) ;

\draw (142,150.5) node   [align=left] {\begin{minipage}[lt]{191.76pt}\setlength\topsep{0pt}
{\fontfamily{ptm}\selectfont {\footnotesize Consider the following \textbf{\textcolor[rgb]{0.29,0.56,0.89}{premise}} and \textbf{\textcolor[rgb]{0.96,0.65,0.14}{hypothesis}} pair. We want to determine if the given statement can be semantically inferred from the given context.}}\\{\fontfamily{ptm}\selectfont {\footnotesize \textbf{\textcolor[rgb]{0.29,0.56,0.89}{Premise}}: \{premise\}}}\\{\fontfamily{ptm}\selectfont {\footnotesize \textbf{\textcolor[rgb]{0.96,0.65,0.14}{Hypothesis}}: \{hypothesis\}}}\\{\fontfamily{ptm}\selectfont {\footnotesize We use a \textcolor[rgb]{0.49,0.83,0.13}{\textbf{commonsense knowledge axiom}} to help determining the entailment relationship between the context and the statement.}}\\{\fontfamily{ptm}\selectfont {\footnotesize \textbf{\textcolor[rgb]{0.49,0.83,0.13}{Commonsense}}: \{commonsense-axiom\}}}\\{\fontfamily{ptm}\selectfont {\footnotesize Rate how helpful this commonsense statement from 1-10 where 1 being not helpful at all and 10 being very helpful, keeping in mind that the gold label for the \textbf{\textcolor[rgb]{0.74,0.06,0.88}{entailment relationship}} is \{inference-label\}. Give a short (maximum 20 tokens) explanation for the rating as well.}}
\end{minipage}};
\draw (3.5,140.25) node  [rotate=-270] [align=left] {\begin{minipage}[lt]{42.84pt}\setlength\topsep{0pt}
{\fontfamily{ptm}\selectfont {\footnotesize \textbf{Instruction}}}
\end{minipage}};

\end{tikzpicture}
    \vspace*{-.1in}

    \caption{\(J_h\) prompt format: We provide the instructions with the test premise-hypothesis pair, a commonsense axiom and gold inference label to get the helpfulness rating of the provided axiom.}
    \label{fig:prompt-helpfulness}
    \vspace{-.15in}
\end{figure}

\newpage
\section{Threshold Selection Experiment} \label{sec:appendix-acc-variation}
The variation of the overall accuracy and the individual inference classes with the access to \(P1\)-generated commonsense axioms, controlled by a threshold for helpfulness rating.

\vspace{-1em} 

\begin{center}
\begin{minipage}{\textwidth}
\centering
\includegraphics[width=\textwidth]{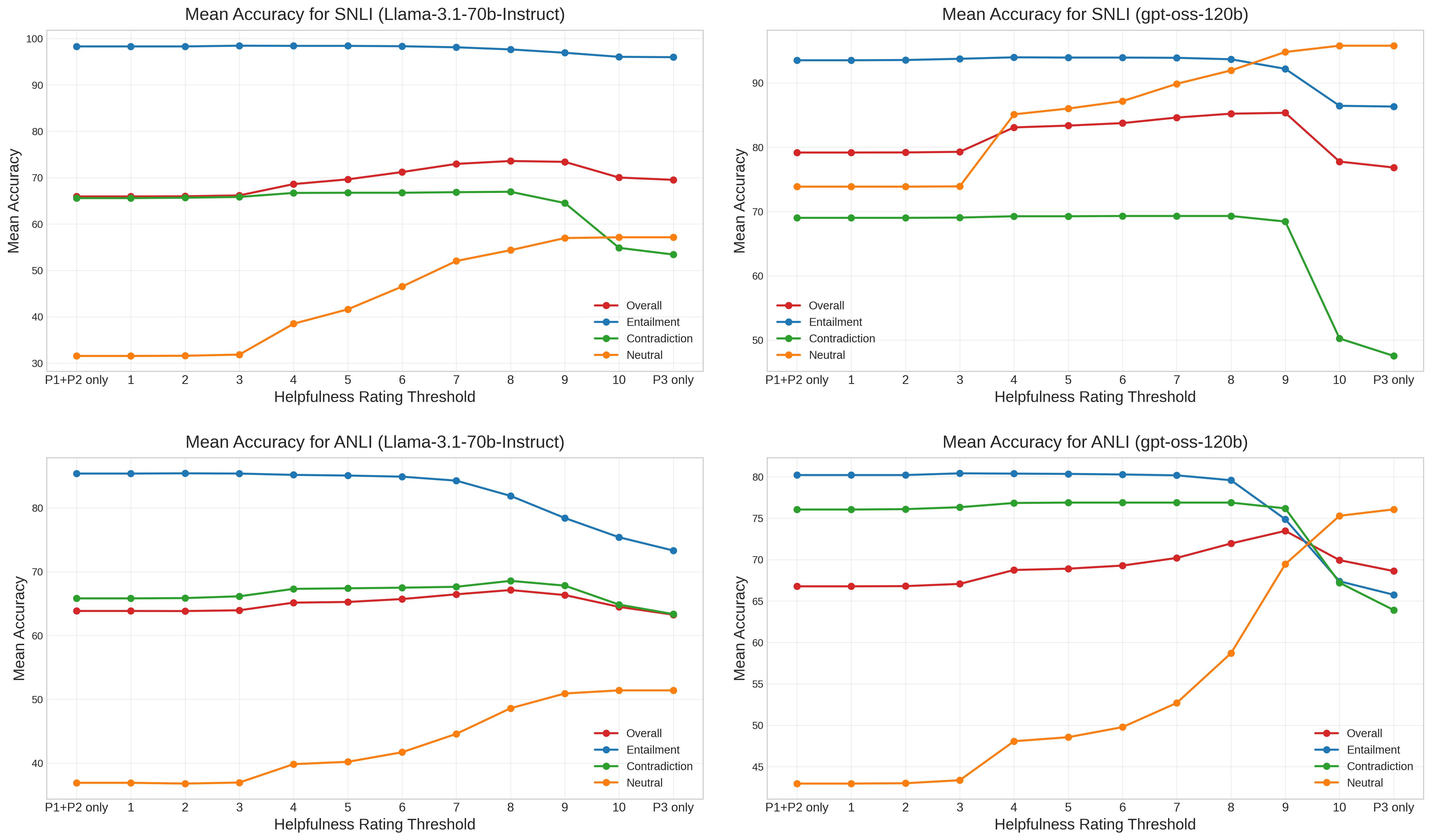}
\captionof{figure}{The variation in accuracy for each inference label and overall, with access to \(P1\)-generated commonsense axioms before prediction, is controlled using a threshold value for helpfulness ratings.}
\end{minipage}
\end{center}





\end{document}